\pdfoutput=1

\documentclass[11pt]{article}
\usepackage[table,xcdraw]{xcolor}
\usepackage[]{acl}

\usepackage{times}
\usepackage{latexsym}
\usepackage{booktabs}
\usepackage{graphicx}
\usepackage{array}
\usepackage{tikz}
\usepackage{bbm}

\usepackage[T1]{fontenc}

\usepackage[utf8]{inputenc}

\usepackage{microtype}

\usepackage{pgfplots}
\pgfplotsset{compat = newest}
\usetikzlibrary{positioning, arrows.meta}
\usepgfplotslibrary{fillbetween}

\usepackage{multirow}
\usepackage{tabularx}
\usepackage{amsmath}
\usepackage{amsxtra}
\usepackage{amsfonts}
\usepackage{multirow}

\newcommand{\eat}[1]{} 

\newcommand{\td}[2]{\if*#1\else^{#1}\fi\if*#2\else_{#2}\fi} 

\newcommand\join\Join 

\DeclareSymbolFont{txsymbolsC}{U}{txsyc}{m}{n}
\SetSymbolFont{txsymbolsC}{bold}{U}{txsyc}{bx}{n}
\DeclareFontSubstitution{U}{txsyc}{m}{n}
\DeclareMathSymbol{\ljoin}{\mathrel}{txsymbolsC}{88}
\DeclareMathSymbol{\rjoin}{\mathrel}{txsymbolsC}{89}

\newsavebox\setminusbox
\newlength\setminuslen

\newcolumntype{C}{>{$\displaystyle}c<{$}} 
\newcolumntype{L}{>{$\displaystyle}l<{$}} 
\newcolumntype{R}{>{$\displaystyle}r<{$}} 
\newcolumntype{H}{>{\setbox0=\hbox\bgroup}c<{\egroup}@{}} 

\newcommand{\B}[3]{B\if*#1\else_{#1}\fi(#2,#3)} 
\newcommand{\I}[3]{I\if*#1\else_{#1}\fi(#2,#3)} 

\makeatletter
\def\imod#1{\allowbreak\mkern10mu({\operator@font mod}\,\,#1)}
\makeatother

\newlength\hspaceoflen

\newcommand\vect[1]{{\boldsymbol{#1}}}
\newcommand\va{\vect{a}}
\newcommand\vb{\vect{b}}
\newcommand\vc{\vect{c}}
\newcommand\vd{\vect{d}}
\newcommand\ve{\vect{e}}
\newcommand\vf{\vect{f}}
\newcommand\vg{\vect{g}}
\newcommand\vh{\vect{h}}
\newcommand\vi{\vect{i}}
\newcommand\vj{\vect{j}}
\newcommand\vk{\vect{k}}
\newcommand\vl{\vect{l}}
\newcommand\vm{\vect{m}}
\newcommand\vn{\vect{n}}
\newcommand\vo{\vect{o}}
\newcommand\vp{\vect{p}}
\newcommand\vq{\vect{q}}
\newcommand\vr{\vect{r}}
\newcommand\vs{\vect{s}}
\newcommand\vt{\vect{t}}
\newcommand\vu{\vect{u}}
\newcommand\vv{\vect{v}}
\newcommand\vw{\vect{w}}
\newcommand\vx{\vect{x}}
\newcommand\vy{\vect{y}}
\newcommand\vz{\vect{z}}

\newcommand\mA{\vect{A}}
\newcommand\mB{\vect{B}}
\newcommand\mC{\vect{C}}
\newcommand\mD{\vect{D}}
\newcommand\mE{\vect{E}}
\newcommand\mF{\vect{F}}
\newcommand\mG{\vect{G}}
\newcommand\mH{\vect{H}}
\newcommand\mI{\vect{I}}
\newcommand\mJ{\vect{J}}
\newcommand\mK{\vect{K}}
\newcommand\mL{\vect{L}}
\newcommand\mM{\vect{M}}
\newcommand\mN{\vect{N}}
\newcommand\mO{\vect{O}}
\newcommand\mP{\vect{P}}
\newcommand\mQ{\vect{Q}}
\newcommand\mR{\vect{R}}
\newcommand\mS{\vect{S}}
\newcommand\mT{\vect{T}}
\newcommand\mU{\vect{U}}
\newcommand\mV{\vect{V}}
\newcommand\mW{\vect{W}}
\newcommand\mX{\vect{X}}
\newcommand\mY{\vect{Y}}
\newcommand\mZ{\vect{Z}}

\DeclareMathAlphabet{\mathcal}{OMS}{cmsy}{m}{n}

\newcommand\cE{\mathcal{E}}

\newcommand\cK{\mathcal{K}}

\newcommand\cN{\mathcal{N}}

\newcommand\cR{\mathcal{R}}

\DeclareMathAlphabet\mathbfcal{OMS}{cmsy}{b}{n}

\accentedsymbol\Abar{{\bar A}}
\accentedsymbol\Bbar{{\bar B}}
\accentedsymbol\Cbar{{\bar C}}
\accentedsymbol\Dbar{{\bar D}}
\accentedsymbol\Ebar{{\bar E}}
\accentedsymbol\Fbar{{\bar F}}
\accentedsymbol\Gbar{{\bar G}}
\accentedsymbol\Hbar{{\bar H}}
\accentedsymbol\Ibar{{\bar I}}
\accentedsymbol\Jbar{{\bar J}}
\accentedsymbol\Kbar{{\bar K}}
\accentedsymbol\Lbar{{\bar L}}
\accentedsymbol\Mbar{{\bar M}}
\accentedsymbol\Nbar{{\bar N}}
\accentedsymbol\Obar{{\bar O}}
\accentedsymbol\Pbar{{\bar P}}
\accentedsymbol\Qbar{{\bar Q}}
\accentedsymbol\Rbar{{\bar R}}
\accentedsymbol\Sbar{{\bar S}}
\accentedsymbol\Tbar{{\bar T}}
\accentedsymbol\Ubar{{\bar U}}
\accentedsymbol\Vbar{{\bar V}}
\accentedsymbol\Wbar{{\bar W}}
\accentedsymbol\Xbar{{\bar X}}
\accentedsymbol\Ybar{{\bar Y}}
\accentedsymbol\Zbar{{\bar Z}}

\accentedsymbol\abar{{\bar a}}
\accentedsymbol\bbar{{\bar b}}
\accentedsymbol\cbar{{\bar c}}
\accentedsymbol\dbar{{\bar d}}
\accentedsymbol\ebar{{\bar e}}
\accentedsymbol\fbar{{\bar f}}
\accentedsymbol\gbar{{\bar g}}
\makeatletter
\@ifundefined{hbar}{}{
        \let\hbar\@undefined
}
\makeatother
\accentedsymbol\hbar{{\bar h}}
\accentedsymbol\ibar{{\bar i}}
\accentedsymbol\jbar{{\bar j}}
\accentedsymbol\kbar{{\bar k}}
\accentedsymbol\lbar{{\bar l}}
\accentedsymbol\mbar{{\bar m}}
\accentedsymbol\nbar{{\bar n}}
\makeatletter
\@ifundefined{obar}{}{
        \let\obar\@undefined
}
\makeatother
\accentedsymbol{\obar}{{\bar o}}
\accentedsymbol\pbar{{\bar p}}
\accentedsymbol\qbar{{\bar q}}
\accentedsymbol\rbar{{\bar r}}
\accentedsymbol\sbar{{\bar s}}
\accentedsymbol\tbar{{\bar t}}
\accentedsymbol\ubar{{\bar u}}
\accentedsymbol\vbar{{\bar v}}
\accentedsymbol\wbar{{\bar w}}
\accentedsymbol\xbar{{\bar x}}
\accentedsymbol\ybar{{\bar y}}
\accentedsymbol\zbar{{\bar z}}

\accentedsymbol\mAhat{{\hat\mA}}
\accentedsymbol\mBhat{{\hat\mB}}
\accentedsymbol\mChat{{\hat\mC}}
\accentedsymbol\mDhat{{\hat\mD}}
\accentedsymbol\mEhat{{\hat\mE}}
\accentedsymbol\mFhat{{\hat\mF}}
\accentedsymbol\mGhat{{\hat\mG}}
\accentedsymbol\mHhat{{\hat\mH}}
\accentedsymbol\mIhat{{\hat\mI}}
\accentedsymbol\mJhat{{\hat\mJ}}
\accentedsymbol\mKhat{{\hat\mK}}
\accentedsymbol\mLhat{{\hat\mL}}
\accentedsymbol\mMhat{{\hat\mM}}
\accentedsymbol\mNhat{{\hat\mN}}
\accentedsymbol\mOhat{{\hat\mO}}
\accentedsymbol\mPhat{{\hat\mP}}
\accentedsymbol\mQhat{{\hat\mQ}}
\accentedsymbol\mRhat{{\hat\mR}}
\accentedsymbol\mShat{{\hat\mS}}
\accentedsymbol\mThat{{\hat\mT}}
\accentedsymbol\mUhat{{\hat\mU}}
\accentedsymbol\mVhat{{\hat\mV}}
\accentedsymbol\mWhat{{\hat\mW}}
\accentedsymbol\mXhat{{\hat\mX}}
\accentedsymbol\mYhat{{\hat\mY}}
\accentedsymbol\mZhat{{\hat\mZ}}

\accentedsymbol\vahat{{\hat\va}}
\accentedsymbol\vbhat{{\hat\vb}}
\accentedsymbol\vchat{{\hat\vc}}
\accentedsymbol\vdhat{{\hat\vd}}
\accentedsymbol\vehat{{\hat\ve}}
\accentedsymbol\vfhat{{\hat\vf}}
\accentedsymbol\vghat{{\hat\vg}}
\accentedsymbol\vhhat{{\hat\vh}}
\accentedsymbol\vihat{{\hat\vi}}
\accentedsymbol\vjhat{{\hat\vj}}
\accentedsymbol\vkhat{{\hat\vk}}
\accentedsymbol\vlhat{{\hat\vl}}
\accentedsymbol\vmhat{{\hat\vm}}
\accentedsymbol\vnhat{{\hat\vn}}
\accentedsymbol\vohat{{\hat\vo}}
\accentedsymbol\vphat{{\hat\vp}}
\accentedsymbol\vqhat{{\hat\vq}}
\accentedsymbol\vrhat{{\hat\vr}}
\accentedsymbol\vshat{{\hat\vs}}
\accentedsymbol\vthat{{\hat\vt}}
\accentedsymbol\vuhat{{\hat\vu}}
\accentedsymbol\vvhat{{\hat\vv}}
\accentedsymbol\vwhat{{\hat\vw}}
\accentedsymbol\vxhat{{\hat\vx}}
\accentedsymbol\vyhat{{\hat\vy}}
\accentedsymbol\vzhat{{\hat\vz}}

\RequirePackage[textsize=scriptsize,color=yellow!15,linecolor=red,disable]{todonotes}
\RequirePackage{enumitem}
\setlist[itemize]{nosep,leftmargin=*,labelwidth=0pt}
\setlist[enumerate]{nosep, leftmargin=*}
\setlist[description]{nosep,leftmargin=.8em}

\RequirePackage{amsmath,amsthm,bm,amssymb}
\usepackage{pifont}
\newcommand{\secref}[1]{\S\ref{#1}}%

\makeatletter
\newcommand\footnoteref[1]{\protected@xdef\@thefnmark{\ref{#1}}\@footnotemark}
\makeatother

\newcommand{\method}{\textsc{KGT5}}

\newcommand{\probP}{\text{I\kern-0.15em P}}

\makeatletter
\g@addto@macro{\normalsize}{%
\setlength{\abovedisplayskip}{3pt plus1pt}%
\setlength{\abovedisplayshortskip}{3pt plus1pt}%
\setlength{\belowdisplayskip}{3pt plus1pt}%
\setlength{\belowdisplayshortskip}{3pt plus1pt}}
\makeatother

\title{Sequence-to-Sequence Knowledge Graph Completion and Question Answering}

\author{Apoorv Saxena\\
  Indian Institute of Science \\
  Bangalore \\
  \small \texttt{apoorvsaxena@iisc.ac.in} \\
  \And
  Adrian Kochsiek \\
  University of Mannheim \\
  Germany \\
  \small \texttt{adrian@informatik.}\\      
  \small \texttt{uni-mannheim.de} \\
  \And Rainer Gemulla \\
  University of Mannheim \\
  Germany \\
  \small \texttt{rgemulla@uni-mannheim.de}
   \\ 
  }

\begin{document}
\maketitle
\begin{abstract}
Knowledge graph embedding (KGE) models represent each entity and relation of a knowledge graph (KG) with low-dimensional embedding vectors.
These methods have recently been applied to KG link prediction and question answering over incomplete KGs (KGQA).
KGEs typically create an embedding for each entity in the graph, which results in large model sizes on real-world graphs with millions of entities. 
For downstream tasks these atomic entity representations often need to be integrated into a multi stage pipeline, limiting their utility.
We show that an off-the-shelf encoder-decoder Transformer model can serve as a scalable and versatile KGE model obtaining state-of-the-art results for KG link prediction and incomplete KG question answering.
We achieve this by posing KG link prediction as a sequence-to-sequence task and exchange the triple scoring approach taken by prior KGE methods with autoregressive decoding.
Such a simple but powerful method reduces the model size up to 98\% compared to conventional KGE models while keeping inference time tractable.
After finetuning this model on the task of KGQA over incomplete KGs, our approach outperforms baselines on multiple large-scale datasets without extensive hyperparameter tuning.\footnote{\label{resources-note}Resources are available at \url{ https://github.com/apoorvumang/kgt5}}



\end{abstract}

\begin{figure*}
  \centering
  \includegraphics[width=\textwidth, trim={-1cm 0 -2.5cm 0},clip ]{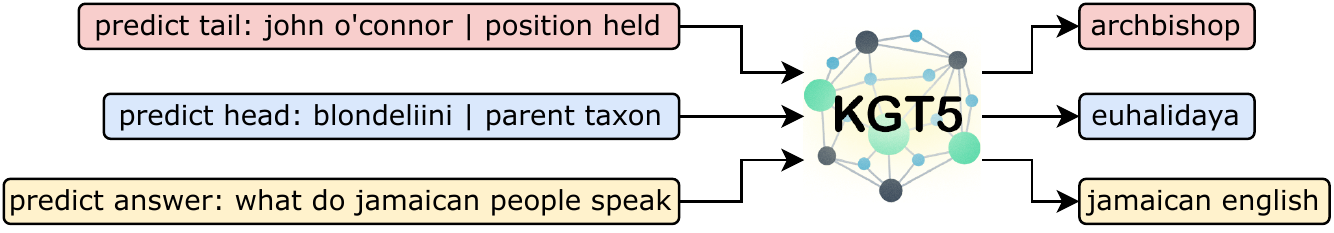}
  \caption{Overview of our method \method{}. KGT5 is first trained on the link prediction task (predicting head/tail entities, given tail/head and relation). For question answering, the same model is further finetuned using QA pairs.
  }
  \label{fig:kgt5-main}
\end{figure*}

\section{Introduction}
\label{sec:introduction}

A knowledge graph (KG) is a multi-relational graph where the nodes are entities from the real world (e.g. \textit{Barack Obama, United States}) and the named edges represent the relationships between them (e.g. \textit{Barack Obama - born in - United States}). 
KGs can be either domain-specific such as WikiMovies \cite{miller2016keyvalue} or public, cross-domain KGs encoding common knowledge such as Wikidata and DBpedia \cite{DBLP:journals/corr/abs-2003-00719}. These graph-structured databases play an important role in knowledge-intensive applications including web search, question answering and recommendation systems \cite{kg_survey_paper}.

Most real-world knowledge graphs are incomplete.
However, some missing facts can be inferred using existing facts in the KG \cite{bordes2013translating}. 
This task termed knowledge graph completion (KGC)\footnote{We use the term KGC for the task of KG link prediction.} has become a popular area of research in recent years \cite{kge_survey} and is often approached using knowledge graph embedding (KGE) models.
KGE models represent each entity and relation of the KG by a dense vector embedding. 
Using these embeddings the model is trained to distinguish correct from incorrect facts.
One of the main downstream applications of KGEs is question answering over incomplete KGs (KGQA) \cite{kge_application_survey}. 

Taking into account the large size of real world KGs (Wikidata contains $\approx$90M entities) and the applicability to downstream tasks, KGE models
should fulfill the following desiderata: 
(i) \emph{scalability} -- i.e. have model size and inference time independent of the number of entities 
(ii) \emph{quality} -- reach good empirical performance
(iii) \emph{versatility} -- be applicable for multiple tasks such as KGC and QA, and 
(iv) \emph{simplicity} -- consist of a single module with a standard architecture and training pipeline.
Traditional KGE models fulfill quality and simplicity. 
They build upon a simple architecture and reach a high quality in terms of KGC.
However, as they create a unique embedding per entity/relation, they scale linearly with the number of entities in the graph, both in model size and inference time, and offer limited versatility.
Methods such as DKRL \cite{Xie_Liu_Jia_Luan_Sun_2016} and KEPLER \cite{wang2021KEPLER} attempt to tackle the scalability issue using compositional embeddings.
However, they fail to achieve quality comparable to conventional KGEs. 
KG-BERT \cite{kg-bert} utilizes pretrained BERT for link prediction and holds potential in terms of versatility as it is applicable to downstream NLP tasks. 
However, it is not scalable due to its underlying cross-encoder.\footnote{
    \citet{time-taken-kgbert} estimate it would take KG-BERT 3 days for an evaluation run on a KG with just 40k entities.
}
QA methods which leverage KGEs outperform traditional KGQA approaches on incomplete KGs, but combining KGEs with the QA pipeline is a non-trivial task;
models that attempt to do this often work on only limited query types (\citealt{huang2019knowledge}; 
\citealt{sun2021faithful}; \citealt{saxena2020improving}) or require multi-stage training and inference pipelines \cite{ren2021lego}.
Here, in order to achieve quality, these models have sacrificed versatility and simplicity.
A comparison of approaches in terms of desiderata is summarized in Tab.~\ref{tab:desiderata} in the appendix.

Our paper shows that all of these desiderata can be fulfilled by a simple sequence-to-sequence (seq2seq) model. To this end, we pose KG link prediction as a seq2seq task and train an encoder-decoder Transformer model \cite{vaswani2017attention} on this task. We then use this model pretrained for link prediction and further finetune it for question answering; while finetuning for QA, we regularize with the link prediction objective. 
This simple but powerful approach, which we call \method{}, is visualised in Fig. \ref{fig:kgt5-main}. 
With such a unified seq2seq approach we achieve 
(i) scalability -- by using compositional entity representations and autoregressive decoding (rather than scoring all entities) for inference
(ii) quality -- we obtain state-of-the-art performance on two tasks 
(iii) versatility -- the same model can be used for both KGC and KGQA on multiple datasets, and 
(iv) simplicity -- we obtain all results using an off-the-shelf model with no task or dataset-specific hyperparameter tuning.
\noindent In summary, we make the following contributions: 
\begin{itemize}
    \item We show that KG link prediction and question answering can be treated as sequence-to-sequence tasks and tackled successfully with a single encoder-decoder Transformer 
    (with the same architecture as T5-small \cite{raffel2020exploring}).
    \item With this simple but powerful approach called \method{}, we reduce model size for KG link prediction up to 98\% while outperforming conventional KGEs on a dataset with 90M entities.
    \item We show the versatility of this approach through the task of KGQA over incomplete graphs. By pretraining on KG link prediction and finetuning on QA, KGT5 performs similar to or better than much more complex methods on multiple large-scale KGQA benchmarks. 
\end{itemize}

\section{Background \& Related Work}
\label{sec:related_work}
Given a set of entities $\cE$ and a set of relations $\cR$, a knowledge graph $\cK \subseteq \cE \times \cR \times \cE$ is a collection of subject-predicate-object $(s,p,o)$ triples.
Link prediction is the task of predicting missing triples in $\cK$ by answering queries of the form of $(s,p,?)$ and $(?,p,o)$. This is typically accomplished using knowledge graph embedding (KGE) models.

Conventional KGEs assign an embedding vector to each entity and relation in the KG. They model the plausibility of $(s,p,o)$ triples via model specific scoring functions $f(e_s, e_p, e_o)$ using the subject ($e_s$), predicate ($e_p$) and object ($e_o$) specific embeddings.
Once trained, these embeddings are used for downstream tasks such as question answering.


Knowledge graph question answering (KGQA) is the task of answering a natural language question using a KG as source of knowledge. 
The questions can be either simple factual questions that require single fact retrieval 
(e.g. \textit{Which languages are spoken in India?}), 
or they can be complex questions that require reasoning over multiple facts in the KG 
(e.g. \textit{What are the genres of movies, in which Leonardo DiCaprio was leading actor?}).
KGEs can be utilized to perform KGQA when the background KGs are incomplete.

In the next few sections we will go into more detail about existing work on KGEs and KGQA.

\subsection{Knowledge Graph Embeddings}
\label{sec:kge}

\textbf{Atomic KGE models.} 
Multiple KGE models have been proposed in the literature, mainly differing in the form of their scoring function $f(e_s, e_p, e_o)$. A comprehensive survey of these models, their scoring functions, training regime and link prediction performance can be found in \citet{kge_survey} and \citet{ruffinelli2020you}.
It is important to note that although these models obtain superior performance in the link prediction task, they suffer from a linear scaling in model size with the number of entities in the KG, and applying them to question answering necessitates separate KGE and QA modules.


\noindent\textbf{Compositional KGE models.}
To combat the linear scaling of the model size with the number of entities in a KG, entity embeddings can be composed of token embeddings.
DKRL~\cite{xie2016representation} embeds entities by combining word embeddings of entity descriptions with a CNN encoder, followed by the TransE scoring function.
KEPLER~\cite{wang2021KEPLER} uses a Transformer-based encoder and combines the typical KGE training objective with a masked language modeling objective.
Both of these approaches encode entities and relations separately which limits the transferability of these models to downstream tasks such as question answering.
MLMLM~\cite{clouatre-etal-2021-mlmlm} encodes the whole query with a RoBERTa-based model and uses \texttt{[MASK]} tokens to generate predictions. However, it performs significantly worse than atomic KGE models on link prediction on large KGs, and is yet to be applied to downstream text-based tasks.

\begin{figure*}
  \centering
  \includegraphics[width=\textwidth, trim={0.2cm 0cm 0.3cm 0cm},clip ]{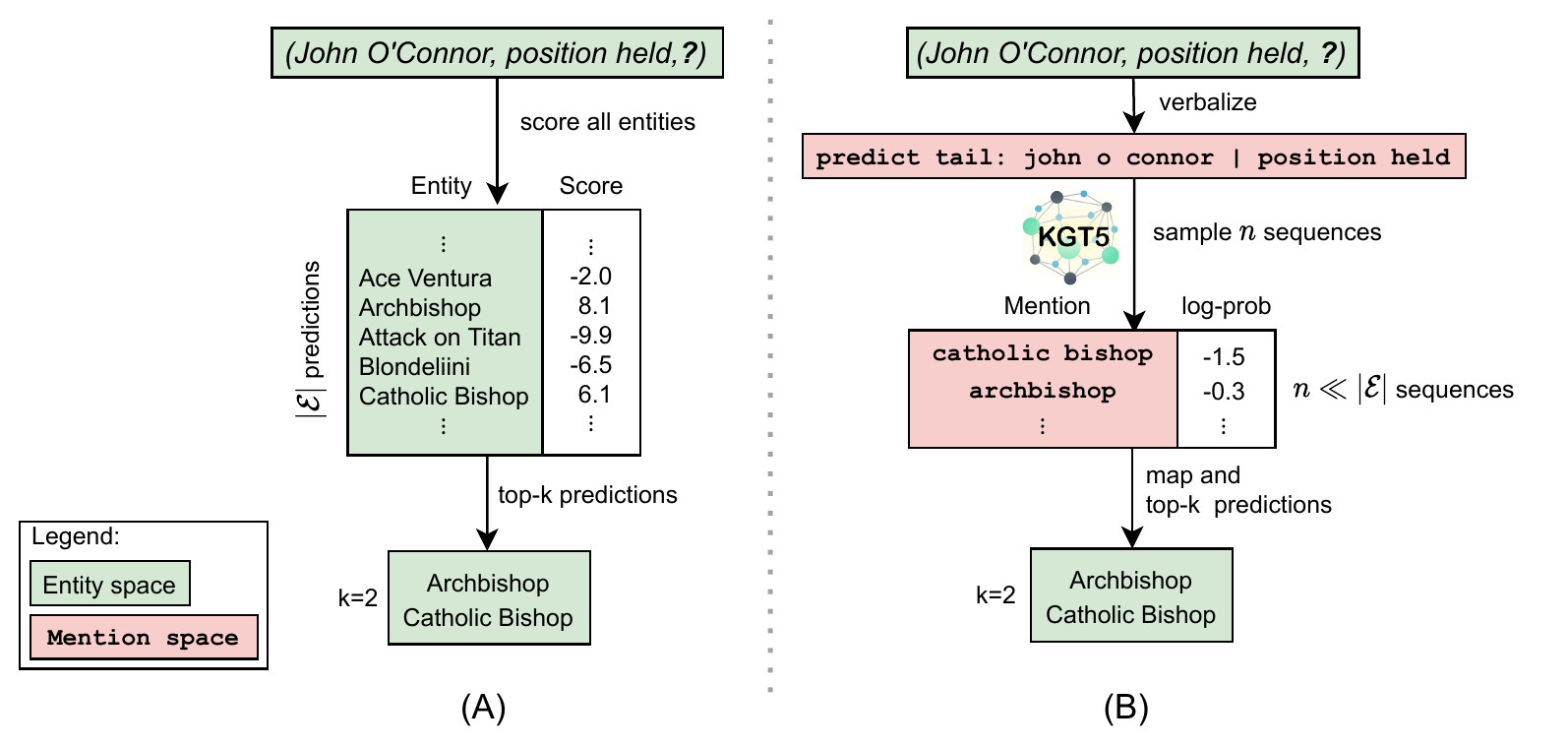}
  \caption{Inference pipeline of (A) conventional KGE models versus (B) \method{} on the link prediction task. Given a query $(s, p, ?)$, we first verbalize it to a textual representation and then input it to the model. A fixed number of sequences are sampled from the model decoder and then mapped back to their entity IDs. This is in contrast to conventional KGEs, where each entity in the KG must be scored. Please see \secref{sec:lp_inference} for more details.
  } 
  \label{fig:kgt5-inference}
\end{figure*}

\subsection{Knowledge Graph Question Answering}
\label{sec:kgqa}

Knowledge Graph Question Answering (KGQA) has been traditionally solved using semantic parsing  (\citealt{Berant2013SemanticPO}; \citealt{bast2015}; \citealt{das2021casebased}) where a natural language (NL) question is converted to a symbolic query over the KG. This is problematic for incomplete KGs, where a single missing link can cause the query to fail.
Recent work has focused on KGQA over incomplete KGs, which is also the focus of our work. These methods attempt to overcome KG incompleteness using KG embeddings (\citealt{huang2019knowledge}; \citealt{saxena2020improving}; \citealt{sun2021faithful}; \citealt{ren2021lego}). In order to use KGEs for KGQA, these methods first train a KGE model on the background KG, and then integrate the learned entity and relation embeddings into the QA pipeline. This fragmented approach brings several disadvantages;
for example \citet{huang2019knowledge}'s method only works for single fact question answering, while EmQL \cite{sun2021faithful} requires prior knowledge of the NL question's query structure. EmbedKGQA \cite{saxena2020improving} is capable of multi-hop question answering but is unable to deal with questions involving more than one entity. Hence, these methods are lacking in versatility. LEGO \cite{ren2021lego} can theoretically answer all first order logic based questions but requires multiple dataset dependent components including entity linking, relation pruning and branch pruning modules; here, to obtain versatility, LEGO has sacrificed simplicity.


\section{The \method{} Model}
\label{sec:model}


We pose both knowledge graph link prediction and question answering as sequence-to-sequence (seq2seq) tasks. We then train a simple encoder-decoder Transformer -- that has the same architecture as T5-small~\cite{raffel2020exploring} but without the pretrained weights -- on these tasks. While training for question answering, we regularize with the link prediction objective. This method, which we call \method{}, results in a scalable KG link prediction model with vastly fewer parameters than conventional KGE models for large KGs. 
This approach also confers simplicity and versatility to the model, whereby it can be easily adapted to KGQA on any dataset regardless of question complexity.

Posing KG link prediction as a seq2seq task requires textual representations of entities and relations, and a verbalization scheme to convert link prediction queries to textual queries; these are detailed in \secref{sec:query_verbalization}.
The link prediction training procedure is explained in \secref{sec:training} and inference in \secref{sec:lp_inference}. The KGQA finetuning and inference pipeline is explained in~\secref{sec:kgqa_training_and_inference}.

\subsection{Textual Representations \& Verbalization}
\label{sec:query_verbalization}
\textbf{Text mapping.} For link prediction we require a one-to-one mapping between an entity/relation and its textual representation.
For Wikidata-based KGs, we use canonical mentions of entities and relations as their textual representation, followed by a disambiguation scheme that appends descriptions and unique ids to the name.\footnote{Please see appendix \ref{sec:text_rep_appendix} for details on textual representations.}
For datasets used for QA only we do not enforce a one-to-one mapping as, in this case, unnecessary disambiguation can even harm model performance.\footnote{This is because QA systems consider surface forms during evaluation, not entity IDs. For example, it will be better to have the same mention for both the single and album version of a song rather than append a unique number to their mentions.}

\noindent\textbf{Verbalization.} We convert $(s,p,?)$ query answering to a sequence-to-sequence task by \textit{verbalizing} the query $(s,p,?)$ to a textual representation. This is similar to the verbalization performed by \citet{lama}, except there is no relation-specific template. For example, given a query \textit{(barack obama, born in, ?)}, we first obtain the textual mentions of the entity and relation and then verbalize it as 
\texttt{'predict tail: barack obama | born in'}. This sequence is input to the model, and output sequence is expected to be the answer to this query, \texttt{'united states'}, which is the unique mention of entity \textit{United States}. 



\subsection{Training \method{} for Link Prediction}
\label{sec:training}
To train KGT5, we need a set of (input, output) sequences. For each triple $(s,p,o)$ in the training graph, we verbalize the queries $(s,p,?)$ and $(?,p,o)$ according to \secref{sec:query_verbalization} to obtain two input sequences. The corresponding output sequences are the text mentions of $o$ and $s$ respectively. KGT5 is trained with teacher forcing \cite{teacher_forcing} and cross entropy loss.\footnote{More details about training are available in Appendix \ref{sec:training_appendix}}

One thing to note is that unlike standard KGE models, we train \textit{without explicit negative sampling}. 
At each step of decoding, the model produces a probability distribution over possible next tokens. While training, this distribution is penalised for being different from the `true' distribution (i.e. a probability of 1 for the true next token, 0 for all other tokens) using cross entropy loss. Hence, this training procedure is most similar to the 1vsAll + CE loss in \citet{ruffinelli2020you}, except instead of scoring the true entity against all other entities, we are scoring the true token against all other tokens at each step, and the process is repeated as many times as the length of the tokenized true entity. This avoids the need for many negatives, and is independent of the number of entities.

\begin{table}[]
\centering
\resizebox{\columnwidth}{!}{%
\begin{tabular}{@{}lrrrr@{}}
\toprule
\multicolumn{1}{c}{\textbf{Dataset}} &
  \multicolumn{1}{c}{\textbf{Entities}} &
  \multicolumn{1}{c}{\textbf{Rels}} &
  \multicolumn{1}{c}{\textbf{Edges}} &
  \multicolumn{1}{c}{\textbf{\begin{tabular}[c]{@{}c@{}}Token.\\ vocab\end{tabular}}} \\ \midrule
WikiKG90Mv2         & 91M  & 1,387 & 601M & 32k \\
Wikidata5M          & 4.8M & 828   & 21M  & 30k \\
MetaQA              & 43k  & 9     & 70k  & 10k \\
WQSP$^\dagger$      & 158k & 816   & 376k & 32k \\
CWQ$^\dagger$ & 3.9M & 326   & 6.9M & 32k \\ \bottomrule
\end{tabular}%
}
\caption{Statistics of the KGs used.~$^{\dagger}$We use subsets of FreeBase~\cite{freebase:datadumps}  for WebQuestionsSP (WQSP) and ComplexWebQuestions (CWQ).
}
\label{tab:kg-stats}
\end{table}

\subsection{Link Prediction Inference}
\label{sec:lp_inference}
In conventional KGE models, we answer a query $(s,p,?)$  by finding the score $f(s,p,o) \; \forall o \in \cE$, where $f$ is the model-specific scoring function. The entities $o$ are then ranked according to the scores.

In our approach, given query $(s,p,?)$, we first verbalize it (\secref{sec:query_verbalization}) before feeding it to KGT5.
We then \textit{sample} a fixed number of sequences from the decoder,\footnote{See Appendix \ref{sec:sampling_appendix} for additional details on sampling and our choice of decoding strategy.} which are then mapped to their entity ids.\footnote{
    The decoded sequence may or may not be an entity mention. We experimented with constrained decoding \cite{decao2020autoregressive} to force the decoder to output only entity mentions; however, we found this unnecessary since the model almost always outputs an entity mention, and increasing the number of samples was enough to solve the issue.
}
By using such a generative model, we are able to approximate (with high confidence) top-$m$ model predictions without having to score all entities in the KG, as is done by conventional KGE models.
For each decoded entity we assign a score equal to the (log) probability of decoding its sequence. This gives us a set of (entity, score) pairs. 
To calculate the final ranking metrics comparable to traditional KGE models, we assign a score of $-\infty$ for all entities not encountered during the sampling procedure. A comparison of inference strategy of conventional KGE models and \method{} is shown in Figure~\ref{fig:kgt5-inference}.


\subsection{KGQA Training and Inference}
\label{sec:kgqa_training_and_inference}
For KGQA, we pretrain the model on the background KG using the link prediction task (\secref{sec:training}). This pretraining strategy is analogous to `KGE module training' used in other KGQA works (\citealt{sun2021faithful}; \citealt{ren2021lego}). The same model is then finetuned for question answering.
Hereby, we employ the same strategy as \citet{roberts-etal-2020-much}: we concatenate a new task prefix (\texttt{predict answer:}) with the input question and define the mention string of the answer entity as output. 
This unified approach allows us to apply \method{} to any KGQA dataset regardless of question complexity, and without the need for sub-modules such as entity linking.

To combat overfitting during QA finetuning (especially on tasks with small KGs) we devise a regularisation scheme: we add link prediction sequences sampled randomly from the background KG to each batch such that a batch consists of an equal number of QA and link prediction sequences.
For inference, we use beam search followed by neighbourhood-based reranking (\secref{sec:setup}) to obtain the model's prediction which is a single answer.

\begin{table*}[ht!]
\centering

\begin{tabular}{@{}lrrrrr@{}}
\toprule
\textbf{Model} & \textbf{MRR}   & \textbf{Hits@1} & \textbf{Hits@3} & \textbf{Hits@10} & \textbf{Params} \\ \midrule
TransE \cite{bordes2013translating} $^\dagger$        & 0.253          & 0.170            & 0.311           & 0.392            & 2,400M            \\
DistMult \cite{yang2015embedding} $^\dagger$      & 0.253          & 0.209           & 0.278           & 0.334            & 2,400M            \\
SimplE \cite{kazemi2018simple} $^\dagger$         & 0.296          & 0.252  & 0.317           & 0.377            & 2,400M            \\
RotatE \cite{sun2018rotate} $^\dagger$         & 0.290           & 0.234           & 0.322           & 0.390             & 2,400M            \\
QuatE \cite{zhang2019quaternion} $^\dagger$         & 0.276          & 0.227           & 0.301           & 0.359            & 2,400M            \\
ComplEx \cite{trouillon2016complex} $^\$$       & \textbf{0.308} & \textbf{0.255}           & -  & \textbf{0.398}   & 614M            \\ \midrule
\method{} (Our method)           & \textbf{0.300} & \textbf{0.267}   & \textbf{0.318}  & \textbf{0.365}   & 60M             \\
ComplEx 14-dim $^\ddagger$  & 0.201          & 0.161           & 0.211           & 0.275            & 67M             \\
ComplEx 26-dim $^\ddagger$  & 0.239          & 0.187           & 0.261           & 0.342            & 125M            \\
KEPLER \cite{wang2021KEPLER} $^{\dagger\dagger}$         & 0.210           & 0.173           & 0.224           & 0.277            & 125M            \\ 
DKRL \cite{Xie_Liu_Jia_Luan_Sun_2016} $^{\dagger\dagger}$         & 0.160           & 0.120           & 0.181           & 0.229            & 20M            \\ 
MLMLM~\cite{clouatre-etal-2021-mlmlm} $^{\ddagger\ddagger}$    &   0.223   &   0.201   &   0.232   &   0.264   &   355M   \\ \midrule

\method{}-ComplEx Ensemble & \textbf{0.336} & \textbf{0.286}  & \textbf{0.362}  & \textbf{0.426}   & 674M            \\ \bottomrule
\end{tabular}%

\caption{Link prediction results on Wikidata5M 
. $\dagger$ results are from the best pre-trained models made available by Graphvite \cite{zhu2019graphvite} . $\ddagger$ results were obtained through a hyperparameter search with LibKGE \cite{libkge}. $\$$ results are from~\cite{kochsiek2021parallel}. $\dagger\dagger$ results are from \citet{wang2021KEPLER}. $\ddagger\ddagger$ results are from \citet{clouatre-etal-2021-mlmlm}. For more details, please see \secref{sec:experiment_kgc}.
}

\label{tab:main-results-kgc}
\end{table*}

\begin{table}[t!]
\centering
\resizebox{\columnwidth}{!}{%
\begin{tabular}{@{}lrrr@{}}
\toprule
\multicolumn{1}{c}{\textbf{Model}} &
  \multicolumn{1}{c}{\textbf{\begin{tabular}[c]{@{}c@{}}Test\\ MRR\end{tabular}}} &
  \multicolumn{1}{c}{\textbf{\begin{tabular}[c]{@{}c@{}}Valid\\ MRR\end{tabular}}} &
  \multicolumn{1}{c}{\textbf{Params}} \\ \midrule
TransE-Concat   & \textbf{0.176} & 0.206          & 18.2B \\
ComplEx-Concat  & \textbf{0.176} & 0.205          & 18.2B \\
ComplEx-MPNet   & 0.099          & 0.126          & 307K  \\
ComplEx & 0.098          & 0.115          & 18.2B \\
TransE-MPNet    & 0.086          & 0.113          & 307K  \\
TransE  & 0.082          & 0.110          & 18.2B \\
\method{} (Our method)             & -\footnoteref{ogb-lsc-note}              & \textbf{0.221} & 60M   \\ \bottomrule
\end{tabular}%
}
\caption{Link prediction results on WikiKG90Mv2. Baseline numbers are from the official leaderboard of OGB-LSC \cite{hu2021ogblsc}. For more details, please see \secref{sec:experiment_kgc}.}
\label{tab:kgc-wikikg90mv2}
\end{table}

\section{Experimental Study}
\label{sec:experimental_study}
We investigate whether KGT5--i.e. a simple seq2seq Transformer model--can be jointly trained to perform both knowledge graph link prediction as well as question answering. Hereby, we first describe the used datasets (\secref{sec:datasets}), the baselines we compared to (\secref{sec:baselines}) and the experimental setup (\secref{sec:setup}). The results of our experiments are analysed in \secref{sec:experiment_kgc}-\secref{sec:limitations}.
Before going into detail, we summarize our key findings:
\begin{enumerate}
    \item For link prediction on large KGs, the text-based approach of \method{} reduces model size to comparable KGE models by up to 98\% and reaches or outperforms current state-of-the-art. 
    \item On the task of KGQA over incomplete KGs, our simple seq2seq approach obtains better results than the current state-of-the-art across multiple datasets.
    \item KG link prediction training might be more beneficial than language modeling pretraining on knowledge intensive tasks such as KGQA.
    \item Although \method{} is good at generalizing to unseen facts, it is rather poor at memorizing facts.
    This problem can be alleviated, if needed, by using an ensemble of \method{} and conventional link prediction or KGQA systems.

\end{enumerate}


\subsection{Datasets}
\label{sec:datasets}
We evaluate the link prediction capability of \method{} on Wikidata5M \cite{wang2021KEPLER} and WikiKG90Mv2 \cite{hu2021ogblsc}, two of the largest publicly available benchmark KGs. 
Although \method{} is designed for large problems, we evaluate on the smaller benchmark KGs FB15k-237~\cite{toutanova2015observed}, WN18RR~\cite{dettmers2018convolutional} and YAGO3-10~\cite{dettmers2018convolutional} for comparability.

We evaluate the QA capabilities of \method{} on three large-scale KGQA benchmark datasets: MetaQA \cite{zhang2017variational}, WebQuestionsSP (WQSP) \cite{yih-etal-2016-value} and ComplexWebQuestions (CWQ) \cite{Talmor2018TheWA}. Questions in MetaQA span from 1-hop to 3-hop questions requiring path-based reasoning on a KG based on WikiMovies \cite{miller2016keyvalue}. WQSP contains both 1-hop and 2-hop path based questions while CWQ contains questions requiring steps such as compositional, conjunctive, comparative and superlative reasoning. Both WQSP and CWQ can be answered using Freebase \cite{freebase:datadumps} as the background KG. 
We create subsets of Freebase using the scheme proposed by \citet{ren2021lego} which results in KGs that are much smaller than Freebase but can still be used to answer all questions in CWQ and WQSP.

Following prior work \cite{sun2019pullnet} we randomly drop 50\% of edges from all KGs to simulate KG incompleteness. 
This stochasticity causes different works to have different KGs, making it hard to compare results without re-implementing methods. 
\citet{ren2021lego} implemented all comparison methods using their own KG splits which they have not yet published.\footnote{Through private communication with the authors we were able to obtain the same KG split for WQSP.}
Our KG split is available along with our implementation\footnoteref{resources-note} and we encourage further studies to use it. 
We do not re-implement comparison methods but instead report the numbers for our methods and baselines separately.
We also report the accuracy obtained by executing the ground truth SPARQL queries (GT query) for test questions.
GT query serves as an estimate of the hardness of a KG split
and  helps us compare model performance across KG splits.
Note that for training all models, we only use (NL question, answer entity) pairs - \textit{no ground truth query information is used for training}.
Statistics of the KGs used in our experiments are shown in Tab.~\ref{tab:kg-stats}. Statistics of the QA datasets are shown in Tab.~\ref{tab:kgqa-dataset-stats}.

\subsection{Comparison Models}
\label{sec:baselines}
For KG completion on Wikidata5M, we compared with several standard KGE models that have been shown to achieve good performance across multiple datasets \cite{ruffinelli2020you} but with a large number of parameters. Among low-parameter models, we compared to the text based approaches KEPLER~\cite{wang2021KEPLER}, DKRL~\cite{Xie_Liu_Jia_Luan_Sun_2016} and MLMLM~\cite{clouatre-etal-2021-mlmlm}. 
We also consider low-dimensional versions of the state-of-the-art method ComplEx.
For the small benchmark KGs we compared with the currently best performing model NBFNet~\cite{nbfnet}.

\begin{table}[t!]
\centering
\resizebox{7cm}{!} {
\begin{tabular}{@{}lll@{}}
\toprule
\textbf{Model} & \textbf{CWQ}  & \textbf{WQSP} \\ \midrule
GT query       & 25.2   & 56.9          \\
Pullnet        & 26.8 \small (+1.6)         & 47.4 \hfill  \small (-9.5)        \\
EmbedKGQA      & -             & 42.5 \hfill \small (-14.4)        \\
LEGO           & 29.4 \small (+4.2)         & 48.5 \hfill \small (-8.4)        \\ \midrule
GT query       & 24.5          & 56.9           \\
\method{}           & \textbf{34.5 \small(+10.0)} & \textbf{50.5 \small(-6.4)} \\ \bottomrule
\end{tabular}
}
\caption{Hits@1 (gain vs GT query) on ComplexWebQuestions (CWQ) and WebQuestionsSP (WQSP) datasets in the 50\% KG setting. Baseline results are from \citet{ren2021lego}. We use the same KG as used by the baselines for WQSP and a slightly \textit{harder} KG for CWQ. Please see \secref{sec:experiment_kgqa} for more details.}
\label{tab:cwq-wqsp-main-results}
\end{table}

For KGQA, we compared against several methods that have been shown to achieve SOTA on QA over incomplete KGs. These include PullNet~\cite{sun2019pullnet}, EmQL \cite{sun2021faithful}, EmbedKGQA \cite{saxena2020improving} and LEGO \cite{ren2021lego}. Additionally, for the MetaQA datasets, we compared with a relation-path finding baseline, which we call PathPred. This simple method maps a NL question to a relation path using distantly supervised data obtained from QA pairs in the training set.\footnote{Please see Appendix \ref{sec:kg_traversal_on_metaqa} for details of PathPred.}

\subsection{Experimental Setup}
\label{sec:setup}

In all our main experiments we used a model with the same architecture as T5-small ($\sim$60M parameters) but without the pretrained weights. For tokenizing sequences, we trained a BPE tokenizer using the  SentencePiece~\cite{sentencepiece} library on the verbalised KGs (see Tab.~\ref{tab:kg-stats} for tokenizer statistics).

We used AdaFactor \cite{shazeer2018adafactor} with a learning rate warmup schedule for link prediction training, batch size 320 and 10\% dropout. We adopted the same procedure as \citet{roberts-etal-2020-much} for QA finetuning - we halved the batch size and fixed the learning rate to 0.001. All experiments were performed using 4 Nvidia 1080Ti GPUs and models were implemented using the HuggingFace~library~\cite{huggingface}. \textit{We performed no dataset-specific hyperparameter tuning} for \method{} and used the same architecture, batch size, dropout and learning rate schedule throughout all experiments.\footnote{The vocabulary size for MetaQA is 10k, compared to $\sim$30k for other datasets. This was necessary in order to train a BPE tokenizer on such a small KG.}
All models were trained until validation accuracy did not significantly increase for 10k steps.\footnote{$\sim$5M steps for large KGs (WD5M, W90M), $\sim$500k steps for smaller KGs and $\sim$30k steps for QA finetuning}

For inference, we used sampling size = 500 for link prediction and beam size = 4 for KGQA. We further performed a neighbourhood-based reranking for KGQA: given question $q$, topic entity from question $e$, predicted answer entity $a$ and (log) probability of predicted entity $p_a$, we compute score for $a$ being answer as
\begin{equation}
\begin{aligned}
    score(a) &= p_a + \alpha \quad \mathrm{if} \  a \in \cN(e)  \\
     &= p_a \qquad \quad \mathrm{otherwise}
\end{aligned}
\end{equation}
where $\alpha$ is a constant hyperparameter and $\cN(e)$ is the $n$-hop neighbourhood of the topic entity ($n$ = 1, 2 or 3). 
Re-ranking was only done on datasets where topic entity annotation is available as part of test questions.



{
  \addtolength{\tabcolsep}{-2.9pt}
\begin{table}[t!]
\centering
\resizebox{\columnwidth}{!}{%
\begin{tabular}{@{}llll@{}}
\toprule
\textbf{Model}           & \textbf{1-hop} & \textbf{2-hop} & \textbf{3-hop} \\ \midrule
GT query                 & 63.3           & 45.8           & 45.3           \\
PullNet                  & 65.1 \footnotesize(+1.8)           & 52.1 \footnotesize(+6.3)           & 59.7 \footnotesize(+14.4)           \\
EmbedKGQA                & \textbf{70.6 \footnotesize(+7.3)}           & 54.3 \footnotesize(+8.5)           & 53.5 \footnotesize(+8.2)           \\
EmQL                     & 63.8 \footnotesize(+0.5)           & 47.6 \footnotesize(+1.8)           & 48.1 \footnotesize(+2.8)           \\
LEGO                     & 69.3 \footnotesize(+6.0)           &\textbf{57.8 \footnotesize(+12.0)}          & 63.8 \footnotesize(+18.5)           \\ \midrule
GT query     & 67.7           & 48.7           & 44.4           \\
PathPred     & 67.7 \footnotesize(+0.0)           & 48.7 \footnotesize(+0.0)           & 44.4 \footnotesize(+0.0)           \\
KGT5         & \textbf{75.0 \footnotesize(+7.3)}          & 36.2 \footnotesize(-8.2)          & \textbf{64.4 \footnotesize(+20.0) }          \\ \midrule
KGT5-PP-Ens.    & \textbf{76.0 \footnotesize (+8.3)}           & \textbf{65.4 \footnotesize (+16.7)}           & \textbf{76.6 \footnotesize (+32.2)}           \\ \bottomrule
\end{tabular}%
}
\caption{
Hits@1 (gain vs GT query) on MetaQA in the 50\% KG setting. Baseline results are from \citet{ren2021lego}. There are two ground truth query (GT query) rows since 
the KG used by baseline models is different from ours. KGT5-PP-Ens. is the KGT5-PathPred ensemble model. Please see \secref{sec:experiment_kgqa} for more details.
}
\label{tab:metaqa-main-results}
\end{table}
}



\subsection{Link Prediction with \method{}}
\label{sec:experiment_kgc}

Tab.~\ref{tab:kgc-wikikg90mv2} shows link prediction performance on WikiKG90Mv2, one of the largest benchmark KGs available. Here we compare against TransE, ComplEx and their variants. *-MPNet and *-concat methods use text embeddings as part of entity representations, and operate on the same textual data as \method{}. \method{} achieves the highest MRR on validation set while having 98\% fewer parameters than the next best performing model on the leaderboard.\footnote{\label{ogb-lsc-note}The authors of OGB-LSC did not provide us with scores on the hidden test set because we used the entity mentions that were provided with the dataset. These entity mentions have now been removed; we provide them for reproducibility on \href{ https://github.com/apoorvumang/kgt5}{our resource website}.}

Tab.~\ref{tab:main-results-kgc} shows link prediction performance on Wikidata5M, a smaller but better studied KG. 
We see that \method{} outperformed all low-parameter count models on all metrics. 
When compared to the large ComplEx model, there is a drop of 0.008 points in MRR and a gain of 0.012 points in hits@1.

We performed a more fine-grained analysis of model predictions according to the type of query for Wikidata5M (Tab.~\ref{tab:kgc-filtering} in the appendix). 
We found that \method{} excelled at answering queries which have none or only a few correct answers in the train set; performance dropped when several entities can be correct for a query. 
This could be due to the nature of sampling: low probability sequences are harder to sample and also harder to rank correctly.
Additionally, the limited sampling (\secref{sec:lp_inference}) may not even provide the correct answer if there exist more known positives than sampled answers.

Based on these observations we created an ensemble of ComplEx and \method{} which answers queries as follows:
if the query does not have answers in the train KG, use \method{};
otherwise use ComplEx (614M).
As shown in Tab.~\ref{tab:main-results-kgc}, the ensemble created by this simple rule outperformed all other single models and achieved the state-of-the-art on Wikidata5M.\footnote{
    In this ensemble \method{} was used to answer 42\% of the queries; the rest were answered by ComplEx
    }$^,$\footnote{
    To the best of our knowledge current state-of-the-art on Wikidata5M is ComplEx published with \citet{kochsiek2021parallel} presented in Tab.~\ref{tab:main-results-kgc}.
}
Such an ensemble neither achieves the goal of scalability nor versatility but instead serves as an ablation to point out weak spots of \method{}.

Tab.~\ref{tab:kgc-small-datasets} in the appendix shows link prediction performance on KGs with $\leq$ 150k entities. Here \method{} sometimes falls behind the baselines; Transformer models are known to struggle when data is scarce, and this could be the reason for poor performance on these small datasets.


\subsection{QA over Incomplete KGs with \method{}}
\label{sec:experiment_kgqa}
Due to the lack of public KG splits, we compared KGQA methods using  \textit{gain over ground truth query model}, which is available for both the comparison methods (from \citealt{ren2021lego}) as well as our methods.\footnote{Details about KGs used by us compared to baselines can be seen in Tab.~\ref{tab:kg-traversal-stats}.}  
Tab.~\ref{tab:cwq-wqsp-main-results} shows hits@1 performance on Freebase-based datasets ComplexWebQuestions and WebQuestionsSP. On both datasets, KGT5 outperformed all baselines. The gains were the largest on ComplexWebQuestions which is the hardest dataset in terms of complexity and KG size.

Tab.~\ref{tab:metaqa-main-results} shows hits@1 performance on the MetaQA datasets. 
On MetaQA 1- and 3-hop, \method{} was either equal or better than all baselines (in terms of gain). On MetaQA 2-hop however, the performance was significantly worse compared to the baselines, and even worse than ground truth querying. We did a more fine-grained analysis of the performance of \method{} on different question types (Tab.~\ref{tab:metaqa_analysis_2hop}-\ref{tab:metaqa_analysis_3hop} in the appendix). We found that \method{} performance suffered most on questions where the head and answer entity were of the same type (for e.g. \textit{actor $\rightarrow$ movie $\rightarrow$ actor} questions). 
These question types are absent in the 1-hop and 3-hop datasets. 
When head and answer entities had different types (for e.g. \textit{director $\rightarrow$ movie $\rightarrow$ language} questions), \method{} was able to answer them better than GT query.

To remedy this issue and create a model more faithful towards the knowledge present in the incomplete KG, we devised an ensemble of \method{} with the PathPred baseline. The ensemble works as follows: Given a question $q$, try to answer it using PathPred. If this returns an empty set, use \method{}. This ensemble outperformed all single models on all MetaQA datasets, often by large margins~(Tab.~\ref{tab:metaqa-main-results}).

Additionally, we performed an ablation to study the effect of neighbourhood reranking on KGQA performance (Tab.~\ref{tab:nbhood-reranking-ablation}). We found that reranking gave small but consistent gains on all datasets.
\begin{table}[t!]
\centering
\begin{tabular}{@{}cllll@{}}
\toprule
\multirow{2}{*}{\textbf{Model}} & \multicolumn{3}{c}{\textbf{MetaQA}}                                                                          & \multicolumn{1}{c}{\multirow{2}{*}{\textbf{WQSP}}} \\
                                & \multicolumn{1}{c}{\textbf{1-hop}} & \multicolumn{1}{c}{\textbf{2-hop}} & \multicolumn{1}{c}{\textbf{3-hop}} & \multicolumn{1}{c}{}                               \\ \midrule
\multicolumn{1}{l}{\method{}}        & 75.0                                 & 36.2                               & 64.4                               & 50.5                                               \\
\multicolumn{1}{r}{$-$ reranking} & 73.1                               & 35.8                               & 63.3                               & 47.2                                               \\ \bottomrule
\end{tabular}
\caption{Effect of neighbourhood reranking on KGQA with 50\% KG. The numbers reported are hits@1.}
\label{tab:nbhood-reranking-ablation}
\end{table}

\subsection{Relation to Knowledge Probing}
\label{sec:knowledge-probing}
Knowledge probing works such as LAMA~\cite{lama} aim to answer the following question: can models (e.g. BERT) which are pretrained on \textit{generic text corpora} with a \textit{language modeling objective} be used as knowledge bases? 
In our case, the model has been explicitly trained with the \textit{link prediction objective}, and a knowledge probing experiment would be akin to checking train set performance of link prediction (which is discussed in \secref{sec:limitations}).
Furthermore, we do not claim that \method{} is as general purpose as large LMs, or that it contains generic world knowledge.
Hence we do not perform knowledge probing experiments on datasets such as T-Rex or Google-RE \cite{lama}.

\begin{table}[t!]
\centering
\resizebox{\columnwidth}{!}{%
\begin{tabular}{@{}lrr@{}}
\toprule
\textbf{Method} & \multicolumn{1}{l}{\textbf{WQSP}} & \multicolumn{1}{l}{\textbf{CWQ}} \\ \midrule
T5-small + QA finetuning & 31.3 & 27.1 \\
KGT5 (50\% KG pretraining)            & 50.5 & 34.5 \\
KGT5 (full KG pretraining)            & 56.1 & 36.5 \\
EmbedKGQA                & 66.6 & -    \\
CBR-KGQA~\cite{cbr-kgqa}                 & \textbf{73.1} & \textbf{70.4} \\ \bottomrule
\end{tabular}%
}
\caption{Hits@1 in the full-KG KGQA setting. For details please see \secref{sec:limitations}.}
\label{tab:full-kg-kgqa}
\end{table}
\begin{table}[t!]
\centering
\begin{tabular}{@{}lrrr@{}}
\toprule
\textbf{Model} & \multicolumn{1}{c}{\textbf{Test MRR}} & \multicolumn{1}{c}{\textbf{Train MRR}} & \multicolumn{1}{l}{\textbf{Params}} \\ \midrule
ComplEx        & 0.308                                 & 0.721                                  & 614M                                \\
KGT5           & 0.300                                 & 0.304                                  & 60M                                 \\ \bottomrule
\end{tabular}

\caption{Train vs. test performance on link prediction on Wikidata5M. Please see \secref{sec:limitations} for details.}
\label{tab:train-test-mrr}
\end{table}

\subsection{KG vs LM pretraining}
\label{sec:kg_vs_lm_pretraining}
We analyzed how generic corpora pretraining performed compared to KG link prediction training for the task of KGQA. We compared with T5-small
\cite{raffel2020exploring}, which has the same architecture as \method{} but pretrained on a mixture of tasks, most notably language modeling on
web text. 
From Tab.~\ref{tab:full-kg-kgqa} we see that \method{} vastly outperformed T5-small. This is not surprising: the data for \method{} pretraining was tailored towards the task performed--KGQA--which was not the case for T5-small. However, this shows that it is the link prediction pretraining that is responsible for the excellent KGQA performance of \method{}.

\subsection{Limitations}
\label{sec:limitations}
\noindent\textbf{Full-KG Question Answering.} Tab. \ref{tab:full-kg-kgqa} shows hits@1 performance in the full KG setting. \method{} performance only marginally improves when pretrained on full KG compared to 50\% KG, and lags far behind both EmbedKGQA (a ComplEx-based method) as well as CBR-KGQA (a semantic parsing method that uses (NL-query, SPARQL-query) parallel data). This indicates that although \method{} excels at generalizing to unseen facts, it may not be good at memorizing facts. This is further supported by the \textit{train set} link prediction performance of \method{} (Tab. \ref{tab:train-test-mrr}); although both ComplEx and \method{} have comparable test MRR, train MRR of ComplEx is significantly better. One possible explanation could be that the reduced model capacity of \method{} -- which has only 60M parameters -- does not allow it to memorize facts seen during pretraining, leading to poor train MRR and full-KG KGQA performance. Hence we recommend against using \method{} as a standalone KGQA method, and it should be used only when query-parsing does not yield good results. 

\noindent\textbf{Use of textual mentions.}
Since \method{} requires textual representations for every entity, it cannot be directly applied to all KGs, and is especially unsuitable for KGs that contain CVT nodes as entities (e.g. full Freebase). Also, care must be taken when comparing models that make use of entity names/descriptions with those that do not. In our experiments, we noticed a significant proportion of validation triples in WikiKG90Mv2 required just text processing (eg. \texttt{<Giovanni Bensi, family name, Bensi>}) and we found a few cases of potential data leakage when definitions are used in WN18RR (eg. \texttt{<hylidae - the amphibian family of tree frogs, hypernym, amphibian family>}). However, from a practical perspective, models which can leverage text data could be more advantageous, and one must assess the pros and cons of a technique before applying it. 

\section{Conclusion and Future Work}
We have shown that KG link prediction and question answering can be treated as seq2seq tasks and tackled successfully with a single encoder-decoder Transformer model. 
We did this by training a Transformer model with the same architecture as T5-small on the link prediction task, and then finetuning it on the QA task. This simple but powerful approach, which we call \method{},
performed competitively with the state-of-the-art methods for KG completion on large KGs while using upto 98\% fewer parameters.
On the task of KGQA on incomplete KGs, we found that our unified approach outperformed baselines on multiple large-scale benchmark datasets. 
Additionally, we compared language modeling pretraining with KG link prediction training and found that for knowledge-intensive tasks such as KGQA, link prediction training could be more beneficial.

One promising direction for future exploration would be to see 
 whether KG link prediction training  could be considered as an additional pretraining objective when training large seq2seq models. 
Furthermore, the impact of model size, and whether larger Transformer models can indeed store more relational information should be investigated.


\bibliography{anthology,custom}
\bibliographystyle{acl_natbib}

\appendix



\begin{table}[ht!]
    \centering
    \resizebox{\columnwidth}{!}{%
    \begin{tabular}{@{}lcccc@{}}
        \toprule
         & scalability & quality & versatility & simplicity   \\
         \midrule
         Traditional KGE & & $\checkmark$ & & $\checkmark$ \\
         DKRL & $\checkmark$ & & & $\checkmark$ \\
         KEPLER & $\checkmark$ & & & $\checkmark$ \\
         KG-Bert & & & $\checkmark$ & $\checkmark$ \\
         MLMLM & $\checkmark$ & & $\checkmark$ & $\checkmark$ \\
         \midrule
         KGE based KGQA & & $\checkmark$ & & \\
         \midrule
         KGT5 & $\checkmark$ & $\checkmark$ & $\checkmark$ & $\checkmark$ \\
         \bottomrule
    \end{tabular}
    }
    \caption{Comparison of related work in terms of the desiderata described in \secref{sec:introduction}.}
    \label{tab:desiderata}
\end{table}

\section{Textual representations of entities and relations}
\label{sec:text_rep_appendix}

For Wikidata based datasets we obtain canonical mentions of entities and relations from the corresponding Wikidata page titles (canonical names).
However, multiple entities can have identical canonical mentions;
we disambiguate such entities by appending the name with their 1-line description if available.
In all other cases of identical canonical mentions we extend each mention with a unique id. This results in a one-to-one mapping between entities and their textual representations. For WikiKG90Mv2 we used the entity names and descriptions provided as part of OGB v1.3.2 data dump. For Wikidata5M, these were extracted from a 2019 WikiData dump.

For the Freebase based question answering datasets, such as WQSP and CWQ, we use the \textit{identifier triples}~\cite{freebasetriples} to retrieve mention strings.
In particular, we use the canonical name (in English) connected by the relation type \texttt{/type/object/name}.
Furthermore, we disambiguate similar to the Wikidata based datasets with an alias retrieved via the relation \texttt{/common/topic/alias} or append part of the description \texttt{/common/topic/description} if available.

\begin{table*}[]
\centering
\resizebox{\textwidth}{!}{%
\begin{tabular}{@{}lrrrrrrrrr@{}}
\toprule
\multicolumn{1}{c}{\multirow{2}{*}{\textbf{Model}}} &
  \multicolumn{3}{c}{\textbf{WN18RR}} &
  \multicolumn{3}{c}{\textbf{FB15k-237}} &
  \multicolumn{3}{c}{\textbf{YAGO3-10}} \\
\multicolumn{1}{c}{} &
  \multicolumn{1}{c}{\textbf{MRR}} &
  \multicolumn{1}{c}{\textbf{H@1}} &
  \multicolumn{1}{c}{\textbf{H@10}} &
  \multicolumn{1}{c}{\textbf{MRR}} &
  \multicolumn{1}{c}{\textbf{H@1}} &
  \multicolumn{1}{c}{\textbf{H@10}} &
  \multicolumn{1}{c}{\textbf{MRR}} &
  \multicolumn{1}{c}{\textbf{H@1}} &
  \multicolumn{1}{c}{\textbf{H@10}} \\ \midrule
ComplEx  &
  0.475 &
  0.438 &
  0.547 &
  0.348 &
  0.253 &
  0.536 &
  \textbf{0.551} &
  \textbf{0.476} &
  \textbf{0.682} \\
NBFNet \cite{nbfnet} &
  \textbf{0.551} &
  \textbf{0.497} &
  \textbf{0.666} &
  \textbf{0.415} &
  \textbf{0.321} &
  \textbf{0.599} &
  - &
  - &
  - \\
KGT5 (Our method) &
  0.508 &
  0.487 &
  0.544 &
  0.276 &
  0.210 &
  0.414 &
  0.426 &
  0.368 &
  0.528 \\ \midrule
KGT5-ComplEx Ensemble &
  0.542 &
  \textbf{0.507} &
  0.607 &
  0.343 &
  0.252 &
  0.377 &
  \textbf{0.552} &
  \textbf{0.481} &
  0.680 \\ \bottomrule
\end{tabular}%
}
\caption{Link prediction results on small KGs ($\leq$ 150k entities). \method{} is generally worse than both NBFNet and ComplEx on FB15k-237 and YAGO3-10 datasets. Performance on WN18RR is somewhat better; however a part of this could be due to the use entity definitions (see \secref{sec:limitations}). Please see \secref{sec:experiment_kgc} for more details.}
\label{tab:kgc-small-datasets}
\end{table*}

\section{Teacher forcing}
\label{sec:training_appendix}

At each step of decoding, the model produces a probability distribution over possible next tokens. While training, this distribution is penalised for being different from the `true' distribution (i.e. a probability of 1 for the true next token, 0 for all other tokens) using cross entropy loss. In teacher forcing \cite{teacher_forcing} the target token is used as the next token during decoding.

An entity usually consists of multiple tokens. Consider an input sequence $input$, target entity mention tokenized as $[w_1, w_2,..,w_T]$ and vocabulary $[v_1,v_2,...,v_M]$. Then
\begin{equation*}
\begin{aligned}
    y_{t,c} &= \mathbbm{1}_{c = w_t} \\
    p_{t,c} &= \probP(v_c | input, w_1, w_2, ..., w_{t-1}) \\
    J_t &= - \sum_{c=1}^{M} y_{t,c}\log p_{t,c} \\
    Loss &= \frac{1}{T}\sum_{t=1}^T J_t
\end{aligned}
\end{equation*}
where $\probP$ is the model's output distribution.

\begin{figure}[t!]
  \centering
  \begin{tikzpicture}
\begin{axis}[
	xlabel=Beam size/Sample size,
	ylabel=MRR,
	width=7.5cm,height=7cm,
    legend pos=south east,
    legend cell align={left},
    ymin=0.22,
    ]

\addplot[color=red,mark=x] coordinates {
(2, 0.272)
(5,0.28)
(10,0.279)
(25,0.275)
(50,0.272)
};

\addplot[color=blue,mark=*] coordinates {
(2,0.248)
(5,0.268)
(10,0.277)
(25,0.288)
(50,0.293)
};

\legend{Beam search,Sampling}
\end{axis}
\end{tikzpicture}
  \caption{Link prediction performance on Wikidata5M. Increasing the sample size steadily increases MRR for the sampling strategy; the opposite effect is seen with beam size $\geq$ 5 and beam search.}
  \label{fig:beam-sample-plot}
\end{figure}
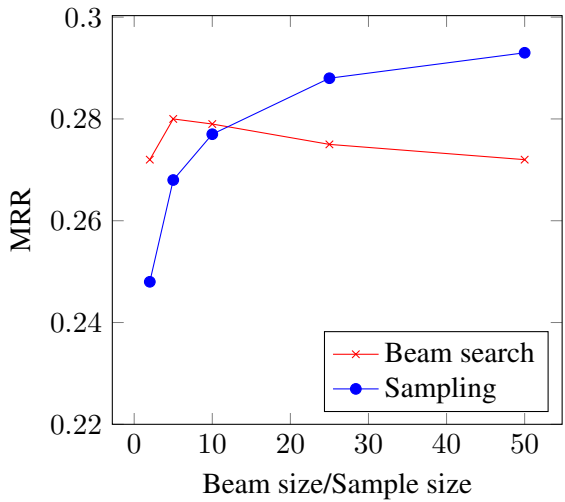

\section{Sampling strategy for link prediction}
\label{sec:sampling_appendix}
At each step of decoding we get a probability distribution over tokens. We sample a token from this distribution and then autoregressively decode until the `stop' token. By repeating this sampling procedure multiple times we can get multiple predictions for the same input sequence. The score for a sequence is the sum of log probabilities for its tokens. For an input sequence $input$, and an entity mention tokenized as $[w_1, w_2, ..., w_T]$, the score for the entity would be
\begin{equation*}
\begin{aligned}
\sum_{t=1}^T \log(\probP(w_t | input, w_1, w_2, ..., w_{t-1}))
\end{aligned}
\end{equation*}
where $\probP$ is the model's output distribution.

Another way to obtain large number predictions could have been beam search \cite{graves_beam_search}. This would also have the advantage of being deterministic and guaranteed to produce as many predictions as we want. Although in theory wider beam sizes should give improved performance, it has been observed that for beam sizes larger than 5, performance of generative models suffers drastically \cite{yang_beam_search_curse} and sampling generally produces better results. We observe the same phenomenon in our work where beam size 50 produces far worse results than sampling 50 times (fig. \ref{fig:beam-sample-plot}). Modifying the stopping criteron \cite{murray_length_bias} or training method \cite{unlikelihood_training} might be helpful solutions that we hope to explore in future work.
\begin{table}[t!]
\centering
\begin{tabular}{@{}llll@{}}
\toprule
\textbf{Dataset} & \textbf{Train} & \textbf{Validation} & \textbf{Test} \\ \midrule
MetaQA 1-hop     & 96,106         & 9,992               & 9,947         \\
MetaQA 2-hop     & 118,980        & 14,872              & 14,872        \\
MetaQA 3-hop     & 114,196        & 14,274              & 14,274        \\
WQSP             & 2,998          & 100                 & 1,639         \\
CWQ              & 27,639         & 3,519               & 3,531         \\ \bottomrule
\end{tabular}
\caption{Numbers of questions in the KGQA datasets used in our experiments.}
\label{tab:kgqa-dataset-stats}
\end{table}
\begin{table}[t!]
\centering
\resizebox{\columnwidth}{!}{%
\begin{tabular}{@{}lllll@{}}
\toprule
Dataset &
  \begin{tabular}[c]{@{}l@{}}Train \\ Questions\end{tabular} &
  \begin{tabular}[c]{@{}l@{}}Distinct \\ Qtypes\end{tabular} &
  \begin{tabular}[c]{@{}l@{}}Distinct \\ NL questions\end{tabular} &
  \begin{tabular}[c]{@{}l@{}}Train \\ QA pairs\end{tabular} \\ \midrule
1-hop & 96,106  & 11 & 161 & 184,884   \\
2-hop & 118,980 & 21 & 210 & 739,782   \\
3-hop & 114,196 & 15 & 150 & 1,521,495 \\ \bottomrule
\end{tabular}
}
\caption{Statistics for MetaQA QA datasets. Since it is a template-based dataset, there is very little linguistic variation - for each linguistic variation, there are more than 1,000 QA pairs on average in the 1-hop dataset. This is further amplified for 2-hop and 3-hop datasets since there are more correct answers on average per question.}
\label{tab:metaqa-traversal-stats}
\end{table}
\begin{table*}[t!]
\centering
\begin{tabular}{c|llll|llll}
\hline
\multirow{3}{*}{Model}       & \multicolumn{4}{c|}{MRR}                                                                                                                                                                       & \multicolumn{4}{c}{Hits@1}                                                                                                                                                                    \\ \cline{2-9} 
                             & \multicolumn{3}{c|}{No. of entities to filter}                                                  & \multicolumn{1}{c|}{\multirow{2}{*}{\begin{tabular}[c]{@{}c@{}}All \\ queries\end{tabular}}} & \multicolumn{3}{c|}{No. of entities to filter}                                                  & \multicolumn{1}{c}{\multirow{2}{*}{\begin{tabular}[c]{@{}c@{}}All \\ queries\end{tabular}}} \\ \cline{2-4} \cline{6-8}
                             & \multicolumn{1}{l|}{0} & \multicolumn{1}{l|}{1 to 10} & \multicolumn{1}{l|}{>10} & \multicolumn{1}{c|}{}                                                                        & \multicolumn{1}{l|}{0} & \multicolumn{1}{l|}{1 to 10} & \multicolumn{1}{l|}{>10} & \multicolumn{1}{c}{}                                                                        \\ \hline
\multicolumn{1}{l|}{ComplEx} & 0.534                   & \textbf{0.351}                             & \textbf{0.045}                     & 0.296                                                                               & 0.464                  & \textbf{0.233}                             & \textbf{0.027}                              & 0.241                                                                              \\
\multicolumn{1}{l|}{\method{}}    & \textbf{0.624}         & 0.215                    & 0.015                              & 0.300                                                                                        & \textbf{0.567}          & 0.164                    & 0.011                     & 0.267                                                                                        \\ \hline
\end{tabular}
\caption{For a test query $(s,r,?)$, there can be multiple entities $o$ such that $(s,r,o)$ is in train set. These entities need to be `filtered' before evaluation. This table shows model performance on queries requiring different amounts of filtering. Dataset is Wikidata5M. The ComplEx checkpoint used in this analysis is slightly worse than the SOTA.}
\label{tab:kgc-filtering}
\end{table*}

\begin{table*}[t!]
\centering
\begin{tabular}{@{}llllll@{}}
\toprule
\multicolumn{1}{c}{\multirow{2}{*}{Model(s)}} & \multicolumn{3}{c}{MetaQA} & \multirow{2}{*}{WQSP} & \multirow{2}{*}{CWQ} \\
\multicolumn{1}{c}{}                          & 1-hop   & 2-hop   & 3-hop  &                       &                      \\ \midrule
Baselines (LEGO, EmbedKGQA, EMQL, PullNet)    & 63.3    & 45.8    & 45.3   & 56.9                  & 25.2                 \\
Ours (\method{}, \method{} Ensemble)                   & 67.7    & 48.7    & 44.4   & 56.9                  & 24.5                 \\ \bottomrule
\end{tabular}
\caption{Percentage of questions answerable using ground truth query. For the baselines that we compare with, we do not have access to the exact same 50\% KG split used by them. This table lists the percentage of questions answerable using GT query, for the KGs used by the comparison models (LEGO, EmbedKGQA, EMQL, PullNet) as well as by our models (\method{}, \method{} + PathPred Ensemble). The GT query numbers for baselines were made available by \citealt{ren2021lego}.} 
\label{tab:kg-traversal-stats}
\end{table*}

\section{Path Predictor on MetaQA}
\label{sec:kg_traversal_on_metaqa}

Being an artificially generated template-based dataset, MetaQA has far more questions than any other dataset that we compare with (Tab.~\ref{tab:kgqa-dataset-stats}). It also has very little variety in the forms of questions (Tab.~\ref{tab:metaqa-traversal-stats}). Hence we try to answer the following question: Can we create a simple model that maps a NL question to a relation path, and then does KG traversal with this path to answer questions? We achieve this by using distant supervision to get the question $\rightarrow$ path mapping data, which is then processed to get the final model. We call this model PathPred.
\textit{We do not use ground truth queries to create this data}.

A question in MetaQA consists of the question text $q_{text}$, a topic entity $h$ and a set of answers $\{a_1, a_2,...\}$ (answers only in train set). Since the topic entity annotation is present for all questions (including test set), we can replace the entity in the question to get a base template $q_{base}$.\footnote{As an example given a $q_{text}$ `\textit{who are the co-actors of Brad Pitt}' and topic entity annotation `\textit{Brad Pitt}', we can get a base template $q_{base}$ as `\textit{who are the co-actors of NE}' where \textit{NE} (named entity) is the substitution string.}

Given a training tuple of $(q_{base}, h, a)$, we find all the k-hop relation paths $[r_1,..,r_k]$ between $h$ and $a$ (k=1,2 or 3 depending on the dataset). We then aggregate these paths for each distinct $q_{base}$, and take the most frequent path as the mapping from $q_{base}$ to relation path. This mapping from question template $q_{base}$ to a relation path $[r_1,..,r_k]$ constitutes the PathPred model.

For a test question $(q_{text}, h)$, we first get $q_{base}$ from $q_{text}$. We then use the aforementioned mapping to get a relation path using $q_{base}$. This relation path is then used to traverse the KG starting from $h$ to arrive at the answer(s).

In the \method{} + PathPred ensemble (\secref{sec:experiment_kgqa}, Tab.~\ref{tab:metaqa-main-results}), we first apply the PathPred technique; if the resulting answer set is empty -- which can happen due to KG incompleteness -- we apply \method{} to get the answer.



\begin{table*}[!htb]
\centering
\resizebox{\textwidth}{!}{%
\begin{minipage}{.5\linewidth}
\begin{tabular}{@{}lrrr@{}}
\toprule
\textbf{Question type} & \multicolumn{1}{l}{\textbf{GTQ}} & \multicolumn{1}{l}{\textbf{KGT5}} & \multicolumn{1}{l}{\textbf{Gain}} \\ \midrule
actor$\rightarrow$movie    & 0.96 & 0.95 & \cellcolor[HTML]{E5F5ED}-0.01 \\
director$\rightarrow$movie & 0.84 & 0.92 & \cellcolor[HTML]{95D5B6}0.08  \\
movie$\rightarrow$actor    & 0.79 & 0.77 & \cellcolor[HTML]{EEF8F3}-0.02 \\
movie$\rightarrow$director & 0.52 & 0.64 & \cellcolor[HTML]{72C69D}0.12  \\
movie$\rightarrow$genre    & 0.48 & 0.63 & \cellcolor[HTML]{57BB8A}0.15  \\
movie$\rightarrow$language & 0.49 & 0.63 & \cellcolor[HTML]{60BF91}0.14  \\
movie$\rightarrow$tags     & 0.72 & 0.7  & \cellcolor[HTML]{EEF8F3}-0.02 \\
movie$\rightarrow$writer   & 0.66 & 0.8  & \cellcolor[HTML]{60BF91}0.14  \\
movie$\rightarrow$year     & 0.46 & 0.45 & \cellcolor[HTML]{E5F5ED}-0.01 \\
tag$\rightarrow$movie      & 1    & 0.96 & \cellcolor[HTML]{F7D7D5}-0.04 \\
writer$\rightarrow$movie   & 0.88 & 0.94 & \cellcolor[HTML]{A7DCC2}0.06  \\ \midrule
All                 & 0.678 & 0.732 & 0.054                           \\ \bottomrule
\end{tabular}%
\end{minipage}%
\begin{minipage}{.6\linewidth}
\begin{tabular}{@{}lrrr@{}}
\toprule
\textbf{Question type} & \multicolumn{1}{l}{\textbf{GTQ}} & \multicolumn{1}{l}{\textbf{KGT5}} & \multicolumn{1}{l}{\textbf{Gain}} \\ \midrule
actor$\rightarrow$movie$\rightarrow$director    & 0.44 & 0.39 & \cellcolor[HTML]{FDF5F4}-0.05 \\
director$\rightarrow$movie$\rightarrow$director & 0.34 & 0.62 & \cellcolor[HTML]{8AD0AE}0.28  \\
director$\rightarrow$movie$\rightarrow$language & 0.37 & 0.77 & \cellcolor[HTML]{57BB8A}0.4   \\
writer$\rightarrow$movie$\rightarrow$writer     & 0.39 & 0.39 & \cellcolor[HTML]{FFFFFF}0     \\
\rowcolor[HTML]{FFFFFF} 
actor$\rightarrow$movie$\rightarrow$genre       & 0.48 & 0.55 & \cellcolor[HTML]{E2F4EB}0.07  \\
\rowcolor[HTML]{FFFFFF} 
director$\rightarrow$movie$\rightarrow$genre    & 0.46 & 0.7  & \cellcolor[HTML]{9BD7B9}0.24  \\
\rowcolor[HTML]{FFFFFF} 
actor$\rightarrow$movie$\rightarrow$actor       & 0.57 & 0.09 & \cellcolor[HTML]{EDA19A}-0.48 \\
\rowcolor[HTML]{FFFFFF} 
writer$\rightarrow$movie$\rightarrow$actor      & 0.51 & 0.31 & \cellcolor[HTML]{F7D7D5}-0.2  \\
\rowcolor[HTML]{FFFFFF} 
actor$\rightarrow$movie$\rightarrow$writer      & 0.48 & 0.44 & \cellcolor[HTML]{FDF7F6}-0.04 \\
\rowcolor[HTML]{FFFFFF} 
movie$\rightarrow$director$\rightarrow$movie    & 0.45 & 0.21 & \cellcolor[HTML]{F6D0CC}-0.24 \\
\rowcolor[HTML]{FFFFFF} 
actor$\rightarrow$movie$\rightarrow$year        & 0.48 & 0.23 & \cellcolor[HTML]{F5CECA}-0.25 \\
\rowcolor[HTML]{FFFFFF} 
writer$\rightarrow$movie$\rightarrow$genre      & 0.4  & 0.59 & \cellcolor[HTML]{B0DFC8}0.19  \\
\rowcolor[HTML]{FFFFFF} 
director$\rightarrow$movie$\rightarrow$actor    & 0.51 & 0.5  & \cellcolor[HTML]{FEFDFC}-0.01 \\
\rowcolor[HTML]{FFFFFF} 
movie$\rightarrow$actor$\rightarrow$movie       & 0.73 & 0.06 & \cellcolor[HTML]{E67C73}-0.67 \\
\rowcolor[HTML]{FFFFFF} 
writer$\rightarrow$movie$\rightarrow$year       & 0.37 & 0.35 & \cellcolor[HTML]{FEFBFA}-0.02 \\
\rowcolor[HTML]{FFFFFF} 
director$\rightarrow$movie$\rightarrow$year     & 0.45 & 0.51 & \cellcolor[HTML]{E6F5EE}0.06  \\
\rowcolor[HTML]{FFFFFF} 
director$\rightarrow$movie$\rightarrow$writer   & 0.47 & 0.44 & \cellcolor[HTML]{FDF9F8}-0.03 \\
\rowcolor[HTML]{FFFFFF} 
movie$\rightarrow$writer$\rightarrow$movie      & 0.5  & 0.3  & \cellcolor[HTML]{F7D7D5}-0.2  \\
\rowcolor[HTML]{FFFFFF} 
writer$\rightarrow$movie$\rightarrow$director   & 0.33 & 0.31 & \cellcolor[HTML]{FEFBFA}-0.02 \\
writer$\rightarrow$movie$\rightarrow$language   & 0.32 & 0.66 & \cellcolor[HTML]{71C69C}0.34  \\
actor$\rightarrow$movie$\rightarrow$language    & 0.4  & 0.54 & \cellcolor[HTML]{C5E8D7}0.14  \\ \midrule
All                               & 0.471 & 0.363 & -0.108                      \\ \bottomrule
\end{tabular}

\end{minipage}%
}
\caption{Hits@1 performance on MetaQA 1-hop (left) and 2-hop (right) validation dataset, 50\% KG setting. GTQ refers to ground truth querying.}
\label{tab:metaqa_analysis_2hop}
\end{table*}



\begin{table*}[ht!]
\centering
\begin{tabular}{@{}lrrr@{}}
\toprule
\textbf{Question type} & \multicolumn{1}{l}{\textbf{GTQ}} & \multicolumn{1}{l}{\textbf{KGT5}} & \multicolumn{1}{l}{\textbf{Gain}} \\ \midrule
movie$\rightarrow$director$\rightarrow$movie$\rightarrow$language & 0.17 & 0.85 & \cellcolor[HTML]{57BB8A}0.68  \\
movie$\rightarrow$director$\rightarrow$movie$\rightarrow$actor    & 0.37 & 0.54 & \cellcolor[HTML]{D5EEE2}0.17  \\
movie$\rightarrow$actor$\rightarrow$movie$\rightarrow$language    & 0.29 & 0.8  & \cellcolor[HTML]{81CCA8}0.51  \\
movie$\rightarrow$writer$\rightarrow$movie$\rightarrow$year       & 0.31 & 0.47 & \cellcolor[HTML]{D8EFE4}0.16  \\
movie$\rightarrow$actor$\rightarrow$movie$\rightarrow$director    & 0.65 & 0.57 & \cellcolor[HTML]{F7D7D5}-0.08 \\
movie$\rightarrow$director$\rightarrow$movie$\rightarrow$genre    & 0.37 & 0.82 & \cellcolor[HTML]{90D3B2}0.45  \\
movie$\rightarrow$writer$\rightarrow$movie$\rightarrow$director   & 0.4  & 0.52 & \cellcolor[HTML]{E2F3EB}0.12  \\
movie$\rightarrow$actor$\rightarrow$movie$\rightarrow$year        & 0.63 & 0.72 & \cellcolor[HTML]{E9F7F0}0.09  \\
movie$\rightarrow$actor$\rightarrow$movie$\rightarrow$writer      & 0.63 & 0.51 & \cellcolor[HTML]{F7D7D5}-0.12 \\
movie$\rightarrow$actor$\rightarrow$movie$\rightarrow$genre       & 0.65 & 0.83 & \cellcolor[HTML]{D3EEE1}0.18  \\
movie$\rightarrow$director$\rightarrow$movie$\rightarrow$writer   & 0.39 & 0.55 & \cellcolor[HTML]{D8EFE4}0.16  \\
movie$\rightarrow$writer$\rightarrow$movie$\rightarrow$genre      & 0.42 & 0.75 & \cellcolor[HTML]{AEDEC7}0.33  \\
movie$\rightarrow$writer$\rightarrow$movie$\rightarrow$actor      & 0.41 & 0.43 & \cellcolor[HTML]{FBFDFC}0.02  \\
movie$\rightarrow$director$\rightarrow$movie$\rightarrow$year     & 0.32 & 0.56 & \cellcolor[HTML]{C4E7D6}0.24  \\
movie$\rightarrow$writer$\rightarrow$movie$\rightarrow$language   & 0.27 & 0.74 & \cellcolor[HTML]{8BD0AF}0.47  \\ \midrule
All                                          & 0.443 & 0.634 & 0.191                          \\ \bottomrule
\end{tabular}%
\caption{Hits@1 performance on MetaQA 3-hop validation dataset, 50\% KG setting. GTQ refers to ground truth querying.}
\label{tab:metaqa_analysis_3hop}
\end{table*}

\end{document}